\documentclass[10pt,twocolumn,letterpaper]{article}

\usepackage{cvpr}

\newcommand{\TODO}[1]{\textbf{\color{red}[TODO: #1]}}

\usepackage{cuted}
\usepackage{capt-of}

\renewcommand{\TODO}[1]{}

\definecolor{cvprblue}{rgb}{0.21,0.49,0.74}
\usepackage[
    breaklinks,
    colorlinks,
    allcolors=cvprblue
]{hyperref}

\usepackage{graphicx}
\graphicspath{{figures/}}

\title{
UniBlendNet: Unified Global, Multi-Scale, and Region-Adaptive Modeling for Ambient Lighting Normalization
}

\author{
Jiatao Dai$^{1}$ \quad Wei Dong$^{1*}$ \quad Han Zhou$^{1}$ \quad Chengzhou Tang$^{2}$ \quad Jun Chen$^{1}$\\
$^{1}$McMaster University \quad $^{2}$University of Manitoba\\
{\tt\small daij56@mcmaster.ca, dongw22@mcmaster.ca} \quad
{\tt\small *Corresponding Author}
}

\begin{document}
\maketitle

\begin{strip}
\vspace{-8pt}
\centering
\includegraphics[width=0.9\textwidth]{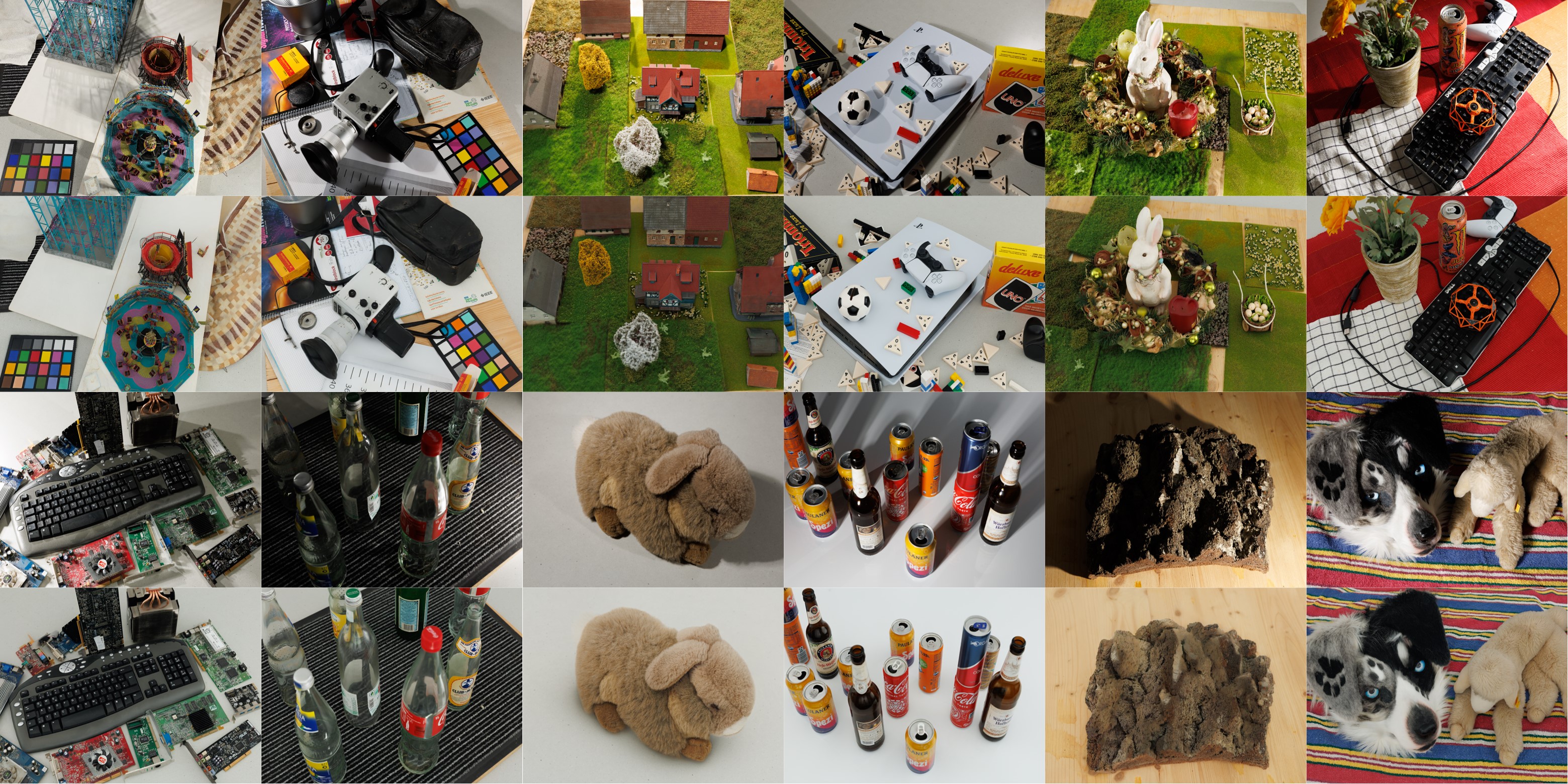}
\captionof{figure}{
Results of UniBlendNet on NTIRE 2026 Ambient Lighting Normalization Challenge~\cite{vasluianu2026ntirealn}. In each image pair, the top image is the input and the bottom image is our corresponding output. Our method achieves top-five performance in terms of PSNR and FID.
}
\label{fig:result}
\vspace{-10pt}
\end{strip}

\begin{abstract}
Ambient Lighting Normalization (ALN) aims to restore images degraded by complex, spatially varying illumination conditions. Existing methods, such as IFBlend, leverage frequency-domain priors to model illumination variations, but still suffer from limited global context modeling and insufficient spatial adaptivity, leading to suboptimal restoration in challenging regions. In this paper, we propose UniBlendNet, a unified framework for ambient lighting normalization that jointly models global illumination, multi-scale structures, and region-adaptive refinement. Specifically, we enhance global illumination understanding by integrating a UniConvNet-based module to capture long-range dependencies. To better handle complex lighting variations, we introduce a Scale-Aware Aggregation Module (SAAM) that performs pyramid-based multi-scale feature aggregation with dynamic reweighting. Furthermore, we design a mask-guided residual refinement mechanism to enable region-adaptive correction, allowing the model to selectively enhance degraded regions while preserving well-exposed areas. This design effectively improves illumination consistency and structural fidelity under complex lighting conditions. Extensive experiments on the NTIRE Ambient Lighting Normalization benchmark demonstrate that UniBlendNet consistently outperforms the baseline IFBlend and achieves improved restoration quality, while producing visually more natural and stable restoration results.
\end{abstract}
\section{Introduction}

Ambient Lighting Normalization (ALN) aims to restore images degraded by complex and spatially varying illuminations. In real-world scenarios, shadows arise from the interaction of multiple light sources, object geometry, and material properties, leading to highly non-uniform illumination distributions. Such degradation not only reduces visual quality but also negatively affects downstream vision tasks (\textit{e.g.}, object detection and semantic segmentation~\cite{dong2024shadowrefiner,le2020fss2sr}).

Recent advances in deep learning have significantly improved shadow removal and illumination restoration performance. Earlier representative deep shadow removal methods mainly focused on shadow detection and removal under relatively constrained conditions~\cite{qu2017deshadownet,wang2018stackedgan,le2019decomposition,ding2019argan}, while more recent methods have achieved stronger restoration ability through improved feature representation and learning strategies, such as masked normalization, global-context modeling, diffusion priors, and region-aware lighting modeling~\cite{jin2021dcshadownet,guo2023shadowformer,guo2023shadowdiffusion,liu2024regionalighting,xu2024dynamicconv}. However, many existing shadow removal benchmarks are still constructed under relatively constrained settings, typically assuming a single dominant light source and limited scene complexity~\cite{hu2021revisitingshadow,vasluianu2023wsrd, dong2024shadowrefiner, rehit}. To address this gap, Vasluianu et al.~\cite{ifblend} introduced the Ambient Lighting Normalization (ALN) task together with the Ambient6K dataset, where multiple light sources, object geometry, and material interactions create significantly more complex illumination degradation. IFBlend further demonstrated the effectiveness of integrating frequency-domain priors for this challenging setting~\cite{ifblend}.

Yet, existing ALN methods still face three limitations. First, complex illumination variations often require scene-level reasoning over long spatial ranges, yet most restoration methods remain dominated by local feature extraction, resulting in insufficient global context modeling. Second, illumination degradation is highly spatially non-uniform, but residual correction is commonly applied in a globally uniform manner, which may over-enhance well-lit regions while under-correcting severely degraded areas. Third, shadows and illumination inconsistencies often appear at different spatial scales, while existing fusion strategies are still limited to adaptively aggregate multi-scale features, reducing robustness under complex lighting conditions.

To address these challenges, we propose UniBlendNet, a unified framework for ALN that explicitly integrates global modeling, multi-scale representation, and region-adaptive refinement. Specifically, we enhance global illumination modeling by incorporating a UniConvNet-based module to capture long-range dependencies~\cite{uniconvnet}. To better handle complex lighting variations, we introduce a Scale-Aware Aggregation Module (SAAM), which performs pyramid-based multi-scale feature aggregation with dynamic reweighting~\cite{sam}. Furthermore, we design a mask-guided residual refinement mechanism, where the predicted guidance mask models the spatial importance of restoration and is used to selectively modulate residual correction, allowing the network to enhance degraded regions while preserving well-exposed areas. These components work in a complementary manner to improve both illumination consistency and structural fidelity.

Extensive experiments on the NTIRE Ambient Lighting Normalization benchmark demonstrate that UniBlendNet consistently outperforms the baseline IFBlend in terms of PSNR and SSIM, while producing visually more natural results, as illustrated in Fig.~\ref{fig:result}. Quantitative improvements over the baseline validate the overall effectiveness of the proposed framework, while the ablation study further confirms the contribution of the three key components, namely global illumination modeling, scale-aware feature aggregation, and region-adaptive residual refinement~\cite{vasluianu2025ntirealn}.

Our contributions can be summarized as follows:
\begin{itemize}
    \item We propose UniBlendNet, a unified framework for ALN that jointly models global illumination, multi-scale structures, and region-adaptive refinement.
    \item We introduce a UniConvNet-based global modeling module and a Scale-Aware Aggregation Module (SAAM) to enhance long-range dependency modeling and multi-scale feature representation.
    \item We design a mask-guided residual refinement mechanism, effectively improving shadow correction while preserving well-lit regions.
\end{itemize}
\section{Related Work}

\subsection{ALN and Shadow Removal}

Shadow removal and illumination normalization have long been fundamental problems in low-level vision. Early approaches primarily relied on hand-crafted priors, region correspondence assumptions, and physical illumination models to estimate shadow regions and recover shadow-free appearance~\cite{finlayson2002removing,finlayson2005removal,finlayson2009entropy,guo2012paired,vicente2014neighbor,zhang2015shadowremover}. While effective under simplified assumptions, these methods often fail in real-world scenarios involving multiple light sources, complex scene geometry, and diverse material properties.

With the development of deep learning, data-driven shadow removal methods have achieved substantial progress by learning direct mappings from shadowed images to shadow-free counterparts. Representative approaches employ convolutional architectures, decomposition strategies, adversarial learning, attention mechanisms, and mask-guided designs to improve restoration quality~\cite{qu2017deshadownet,wang2018stackedgan,le2019decomposition,ding2019argan,he2021maskshadownet,jin2021dcshadownet,liu2021generation2removal,guo2023shadowformer,guo2023shadowdiffusion}. However, many existing benchmarks are constructed under relatively constrained settings, typically assuming a single dominant light source and limited scene complexity~\cite{hu2021revisitingshadow,vasluianu2023wsrd,ifblend,vasluianu2025ntirealn}. As a result, these models often struggle to generalize to more realistic illumination conditions involving self-shadows, multiple cast shadows, and spatially varying lighting distributions~\cite{vasluianu2023wsrd,ifblend,vasluianu2025ntirealn}.

To address these limitations, Vasluianu \textit{et al.}~\cite{ifblend} introduced the Ambient Lighting Normalization (ALN) task along with the Ambient6K dataset, which provides a more realistic benchmark for studying restoration under complex illumination~\cite{ifblend,vasluianu2025ntirealn}. In the same work, IFBlend was proposed as a strong baseline that combines spatial and frequency-domain representations for illumination normalization. By jointly modeling image and frequency information, IFBlend improves the restoration of shadowed regions without requiring explicit shadow-mask priors~\cite{ifblend}.

However, existing methods, including IFBlend, still struggle to jointly model three key aspects: (1) global illumination consistency, (2) multi-scale variation of lighting degradation, and (3) spatially adaptive correction. Specifically, methods based mainly on local convolutional processing are often insufficient for capturing long-range illumination dependencies, while globally shared residual prediction tends to cause over-enhancement in well-lit regions and under-correction in severely degraded shadow regions. In addition, features extracted at a single scale or weakly aggregated across scales are often inadequate for handling the coexistence of smooth shading transitions and sharp shadow boundaries. These limitations indicate that effective ALN requires a unified modeling of global context, multi-scale variation, and region-adaptive restoration~\cite{guo2023shadowformer,liu2024regionalighting,xu2024dynamicconv,hu2024unveiling,dong2024shadowrefiner}. In contrast to prior works that typically emphasize only one or two of these aspects, our method integrates all three within a single ALN framework built on IFBlend.

\subsection{Image Restoration, Multi-Scale Modeling, and Global Context Learning}

Image restoration aims to recover high-quality images from degraded observations and has been widely studied in tasks such as denoising, deblurring, deraining, and dehazing. Recent methods have achieved remarkable performance by leveraging encoder-decoder architectures, residual learning, attention mechanisms, and transformer-based global interaction modules~\cite{ronneberger2015unet,liang2021swinir,liu2021swin,liu2022convnext,zamir2022restormer,dosovitskiy2020vit, sgllie, glare, gppllie, ecmamba, varlide, zhou2025lita}. Among these, multi-scale modeling and global context learning have been shown to be particularly important for handling spatially varying degradations~\cite{nah2017multiscale,liu2019griddehazenet,liu2022griddehazenetplus,qin2020ffanet,song2023vitdehazing}.

To improve multi-scale representation, Yu \textit{et al.}~\cite{sam} proposed a scale-aware design that extracts pyramid features at different resolutions and dynamically fuses them according to image-dependent scale responses. This design enables effective aggregation of aligned multi-scale context while maintaining relatively low computational cost. Such a property is especially beneficial for ALN, where illumination degradation often appears at different spatial scales and requires adaptive feature integration.

On the other hand, effective global context modeling is essential for capturing long-range illumination dependencies. Recent restoration and shadow removal methods have shown that global reasoning can significantly improve illumination consistency and structural restoration~\cite{guo2023shadowformer,lu2024hirformer, dehazedct, dwtffc}. UniConvNet~\cite{uniconvnet} expands the effective receptive field through the aggregation of convolution kernels with progressively increasing sizes, enabling more stable and expressive global feature representation. Compared with conventional local convolution blocks, UniConvNet provides stronger long-range dependency modeling, which is critical for understanding scene-level illumination patterns in ALN.

Motivated by these observations, we propose UniBlendNet, which explicitly integrates global context modeling, multi-scale feature aggregation, and spatially adaptive refinement within a unified framework. Specifically, we incorporate a UniConvNet-based module to enhance global illumination modeling, introduce a Scale-Aware Aggregation Module (SAAM) to improve multi-scale representation, and design a mask-guided residual refinement mechanism to enable spatially adaptive correction. Unlike prior shadow removal and restoration methods that mainly improve one aspect such as global reasoning or multi-scale fusion in isolation, UniBlendNet combines these components with the frequency-aware restoration capability of IFBlend for ALN, leading to more robust and accurate restoration under complex non-uniform illumination conditions.
\section{Method}

\begin{figure*}[t]
    \centering
    \setlength{\abovecaptionskip}{1mm}
    \includegraphics[width=0.9\linewidth]{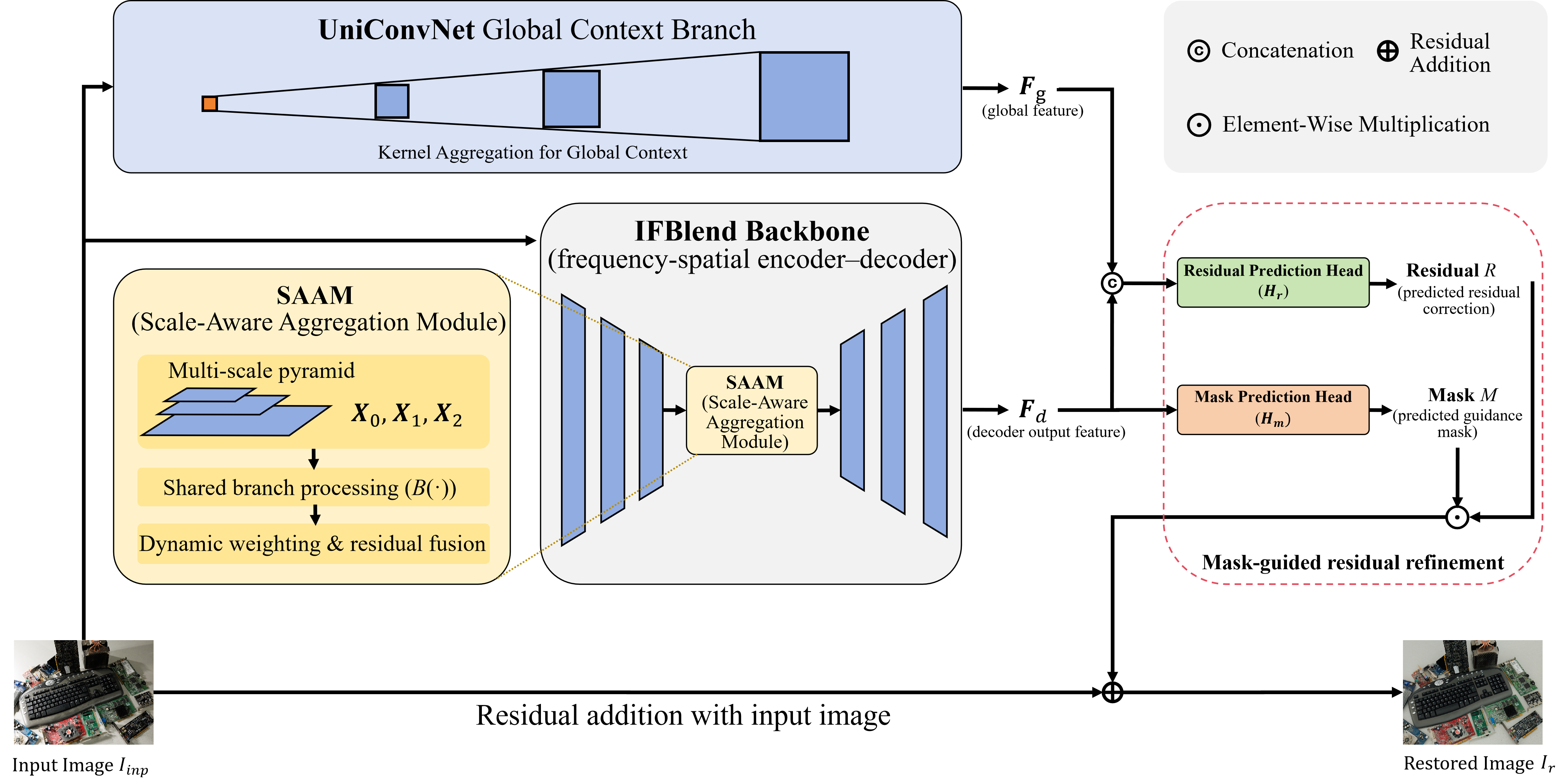}
    \caption{
    Overview of the proposed UniBlendNet for Ambient Lighting Normalization (ALN). 
    Built upon IFBlend, the model introduces a UniConvNet-based global context branch, a Scale-Aware Aggregation Module (SAAM) at the bottleneck, and a mask-guided residual refinement mechanism. 
    The final output is generated by adding the mask-modulated residual correction to the input image.
    }
    \label{fig:pipeline}
    \vspace{-1.5mm}
\end{figure*}

Ambient Lighting Normalization (ALN) aims to restore images degraded by complex and spatially varying illumination conditions~\cite{vasluianu2024aln,vasluianu2025ntirealn}. Built upon IFBlend~\cite{ifblend}, we propose \textbf{UniBlendNet}, which enhances the baseline with three complementary components: a UniConvNet-based global context branch, a scale-aware multi-scale aggregation module, and a mask-guided residual refinement mechanism. As illustrated in Fig.~\ref{fig:pipeline}, these components are integrated into a unified frequency-spatial restoration framework to improve scene-level illumination modeling, multi-scale feature representation, and spatially adaptive correction~\cite{guo2023shadowformer,xiao2024homoformer,xu2024dynamicconv}.

\subsection{Overall Framework}

Given an input image ${\mathbf I}_{inp}$ captured under non-uniform illumination, UniBlendNet aims to produce a restored image ${\mathbf I}_{r}$ with normalized lighting and recovered structural details. The overall architecture follows the encoder-decoder design of IFBlend~\cite{ifblend}, while introducing three key modifications. First, a UniConvNet-based global context branch is incorporated to provide stronger scene-level illumination modeling. Second, a \textbf{Scale-Aware Aggregation Module (SAAM)} is inserted at the bottleneck to enhance multi-scale feature aggregation. Third, a mask-guided residual refinement mechanism is introduced to selectively enhance degraded regions while preserving well-lit areas.

Formally, the restoration process can be expressed as:
\vspace{-1.5mm}
\begin{equation}
{\mathbf I}_{r} = {\mathbf I}_{inp} + {\mathbf M} \odot {\mathbf R},
\label{eq:final_output}
\end{equation}
where ${\mathbf R}$ denotes the predicted residual correction, ${\mathbf M}$ is the predicted soft guidance mask, and $\odot$ denotes element-wise multiplication. As shown in Fig.~\ref{fig:pipeline}, the residual is predicted first and then modulated once by the guidance mask before being added back to the input image. This formulation enables region-adaptive restoration, rather than applying uniform correction to all pixels.

\subsection{IFBlend-based Frequency-Spatial Backbone}

Our method is built upon IFBlend~\cite{ifblend}, which combines spatial and frequency priors for illumination normalization. In our framework, we retain the frequency-spatial restoration backbone proposed in IFBlend, because ALN requires both global illumination adjustment and local detail recovery~\cite{ifblend,vasluianu2024aln}. However, we further enhance IFBlend with stronger global context modeling, more effective multi-scale aggregation, and spatially adaptive residual correction, which are not explicitly addressed before~\cite{guo2023shadowformer,xu2024dynamicconv,liu2024regionalighting}. These designs are motivated by the observation that ALN requires not only frequency-aware local restoration, but also scene-level illumination reasoning and spatially selective correction under highly non-uniform degradation.


Given an input feature map, the encoder progressively extracts hierarchical representations while jointly incorporating RGB guidance and frequency decomposition. At each encoder stage, the feature is processed by a U-Net-style compression block~\cite{ronneberger2015unet} to produce latent spatial features. In parallel, a Discrete Wavelet Transform (DWT)~\cite{mallat1989wavelet} is performed to decompose the feature into low-frequency and high-frequency components:
\vspace{-1.5mm}
\begin{equation}
{\mathbf F}_{lf}, {\mathbf F}_{hf} = \mathrm{DWT}({\mathbf F}),
\label{eq:dwt}
\end{equation}
where ${\mathbf F}_{lf}$ captures coarse illumination and structural information, while ${\mathbf F}_{hf}$ preserves high-frequency texture details. In addition, an RGB branch is introduced to extract complementary appearance cues from the image guidance stream. These features are fused with the encoder representation to form a joint frequency-spatial feature space.

During decoding, the high-frequency features are re-injected into the corresponding decoder stages to recover local textures and sharp transitions. To further improve cross-domain interaction, a weighted attention module is used to fuse RGB features and high-frequency features, allowing the decoder to benefit from both appearance guidance and frequency priors. This frequency-spatial collaborative design provides the strong restoration capability inherited from IFBlend and serves as the foundation of our method.

\subsection{UniConvNet-based Global Context Modeling}

Although IFBlend effectively exploits spatial and frequency information, its original backbone still mainly operates through local feature extraction and hierarchical reconstruction. However, ALN is not purely a local restoration problem: the illumination state of one region is often correlated with broader scene-level lighting patterns, such as large cast shadows, smooth shading transitions, and long-range intensity imbalance across the image~\cite{guo2023shadowformer,xiao2024homoformer}. Therefore, relying only on local receptive fields may lead to inconsistent restoration across distant regions. To address this issue, we enhance global context modeling by incorporating a UniConvNet-based module~\cite{uniconvnet}. The location of this branch in the overall framework is shown in Fig.~\ref{fig:pipeline}.

UniConvNet expands the effective receptive field through the aggregation of convolution kernels with progressively increasing sizes, enabling stronger long-range dependency modeling while maintaining stable feature representation~\cite{uniconvnet}. In our framework, the UniConvNet-based branch is utlized to extract global context features:
\begin{equation}
{\mathbf F}_{g} = \mathcal{G}({\mathbf I}_{inp}),
\label{eq:gcb}
\end{equation}
where $\mathcal{G}(\cdot)$ denotes the UniConvNet-based global context extractor and is parallel to the IFBlend backbone. That is, ${\mathbf F}_{g}$ is directly extracted from the input image and then fused with the final decoder feature before residual prediction, as illustrated in Fig.~\ref{fig:pipeline}. In this way, the global context branch complements the frequency-spatial backbone by providing scene-level illumination cues that are difficult to capture through local encoder-decoder processing alone. This design improves illumination understanding and strengthens restoration consistency across distant regions.


\subsection{Multi-scale Feature Aggregation}

To better model illumination degradation at different spatial scales, we introduce a \textbf{Scale-Aware Aggregation Module (SAAM)} at the bottleneck stage. As shown in Fig.~\ref{fig:pipeline}, this module is inserted between the encoder and decoder to enhance the bottleneck representation before reconstruction. The motivation is that illumination variations in ALN may appear as large smooth shading changes, medium-scale shadow transitions, or fine local details. Therefore, effective restoration requires feature aggregation across multiple spatial scales, rather than relying only on a single-resolution bottleneck representation~\cite{nah2017multiscale,sam}.

Given the bottleneck feature ${\mathbf X} \in \mathbb{R}^{C \times H \times W}$, SAAM first constructs a hierarchical three-level pyramid:
\begin{equation}
{\mathbf X}_{0} = {\mathbf X}, \quad
{\mathbf X}_{1} = \mathrm{Down}_{2}({\mathbf X}), \quad
{\mathbf X}_{2} = \mathrm{Down}_{4}({\mathbf X}),
\label{eq:sam_pyramid}
\end{equation}
where $\mathrm{Down}_{2}$ and $\mathrm{Down}_{4}$ denote bilinear downsampling by factors of $2$ and $4$, respectively. Each pyramid feature is then processed by a shared-weight convolutional branch:
\begin{equation}
{\mathbf Y}_{i} = \mathcal{B}({\mathbf X}_{i}), \quad i \in \{0,1,2\},
\label{eq:sam_branch}
\end{equation}
where $\mathcal{B}(\cdot)$ denotes the shared pyramid branch. The outputs from lower resolutions are then upsampled back to the original feature size.

To adaptively fuse multi-scale features, we compute global descriptors for each branch using global average pooling, concatenate them, and predict scale-wise dynamic weights through a lightweight MLP:
\begin{equation}
[{\mathbf w}_{0}, {\mathbf w}_{1}, {\mathbf w}_{2}] = \mathrm{MLP}([\mathbf{v}_{0}, \mathbf{v}_{1}, \mathbf{v}_{2}]),
\label{eq:sam_weight}
\end{equation}
where $\mathbf{v}_{i}$ denotes the pooled descriptor of ${\mathbf Y}_{i}$. The final SAAM output is obtained by residual fusion:
\begin{equation}
{\mathbf X}_{saam} = {\mathbf X} + {\mathbf w}_{0} \odot {\mathbf Y}_{0}
+ {\mathbf w}_{1} \odot {\mathbf Y}_{1}
+ {\mathbf w}_{2} \odot {\mathbf Y}_{2}.
\label{eq:sam_out}
\end{equation}

This module enables the network to dynamically emphasize the most useful scale for a given input, leading to more robust modeling of complex illumination variations~\cite{nah2017multiscale,sam,Hou_2021_CVPR}.

\subsection{Mask-guided Residual Refinement}

A key limitation of conventional restoration frameworks is that they typically apply residual correction uniformly across the whole image~\cite{xu2024dynamicconv,jie2022mgrln}. In ALN, however, lighting degradation is highly spatially non-uniform: shadow regions require stronger correction, while well-lit regions should remain largely unchanged. To explicitly model this property, we design a mask-guided residual refinement mechanism. As shown in Fig.~\ref{fig:pipeline}, this part contains two separate prediction heads: a \emph{mask prediction head} $\mathcal{H}_{m}(\cdot)$ for estimating the soft guidance mask, and a \emph{residual prediction head} $\mathcal{H}_{r}(\cdot)$ for generating the residual correction.

Given the decoder output feature ${\mathbf F}_{d}$, we first predict a soft guidance mask ${\mathbf M}$ through the mask prediction head:
\begin{equation}
{\mathbf M} = \sigma(\mathcal{H}_{m}({\mathbf F}_{d})),
\label{eq:mask}
\end{equation}
where $\mathcal{H}_{m}(\cdot)$ denotes the mask prediction head and $\sigma(\cdot)$ is the sigmoid activation. Although the mask branch is supervised by a binary pseudo mask, the predicted output ${\mathbf M}$ is continuous in $[0,1]$ and therefore acts as a spatially varying gating map that controls the strength of residual correction at each location.

The residual correction ${\mathbf R}$ is predicted separately from the decoder feature, together with the global context feature when the UniConvNet branch is used:
\vspace{-1mm}
\begin{equation}
{\mathbf R} = \mathcal{H}_{r}([{\mathbf F}_{d}, {\mathbf F}_{g}]),
\label{eq:residual}
\end{equation}
where $[\cdot,\cdot]$ denotes channel-wise feature concatenation. and $\mathcal{H}_{r}(\cdot)$ denotes the residual prediction head. The predicted residual is then modulated once by the guidance mask according to Eq.~(\ref{eq:final_output}), as illustrated in Fig.~\ref{fig:pipeline}.

In this way, the network learns \emph{where} to restore and \emph{how much} to restore. This process can be interpreted as a spatially varying modulation of residual learning~\cite{xu2024dynamicconv,jie2022mgrln}. It effectively reduces over-enhancement in well-exposed regions while improving correction strength in degraded shadow areas~\cite{xu2024dynamicconv,jie2022mgrln,liu2024regionalighting}.

\subsection{Training Objective}

We adopt a multi-objective loss function for joint training:
\begin{equation}
\mathcal{L} = \mathcal{L}_{rec}
+ \alpha_{1}\mathcal{L}_{ssim}
+ \alpha_{2}\mathcal{L}_{grad}
+ \alpha_{3}\mathcal{L}_{perc}
+ \lambda \mathcal{L}_{mask},
\label{eq:total_loss}
\end{equation}
where $\mathcal{L}_{rec}$ denotes the reconstruction loss, $\mathcal{L}_{ssim}$ is the structural similarity loss~\cite{wang2004ssim}, $\mathcal{L}_{grad}$ is the gradient consistency loss, $\mathcal{L}_{perc}$ is the perceptual loss, and $\mathcal{L}_{mask}$ is the mask supervision loss. These losses can be computed as:
\begin{equation}
\mathcal{L}_{rec} = \| {\mathbf I}_{r} - {\mathbf I}_{gt} \|_{1}.
\label{eq:l1}
\end{equation}
\vspace{-4mm}
\begin{equation}
\mathcal{L}_{ssim} = 1 - \mathrm{SSIM}({\mathbf I}_{r}, {\mathbf I}_{gt}).
\label{eq:ssim}
\end{equation}
\vspace{-4mm}
\begin{equation}
\mathcal{L}_{grad} = \| \nabla {\mathbf I}_{r} - \nabla {\mathbf I}_{gt} \|_{1},
\label{eq:grad}
\end{equation}
\vspace{-4mm}
\begin{equation}
\mathcal{L}_{perc} = \| \phi({\mathbf I}_{r}) - \phi({\mathbf I}_{gt}) \|_{1},
\label{eq:perc}
\end{equation}
where $\nabla$ denotes the image gradient operator and $\phi(\cdot)$ is the perceptual feature extractor. ${\mathbf I}_{r}$ and ${\mathbf I}_{gt}$ are restored outputs and clean images, respectively. Moreover, we construct a binary pseudo mask ${\mathbf M}_{gt}$ from the positive relative grayscale difference between degraded and clean images, followed by local averaging and threshold refinement. The predicted soft mask ${\mathbf M}$ is then supervised by an L1 loss:
\vspace{-2mm}
\begin{equation}
\mathcal{L}_{mask} = \|{\mathbf M} - {\mathbf M}_{gt}\|_{1}.
\label{eq:mask_loss}
\end{equation}





\section{Experiments}

\subsection{Experimental Setup}

\begin{figure*}[t]
    \centering
    \includegraphics[width=0.8\linewidth]{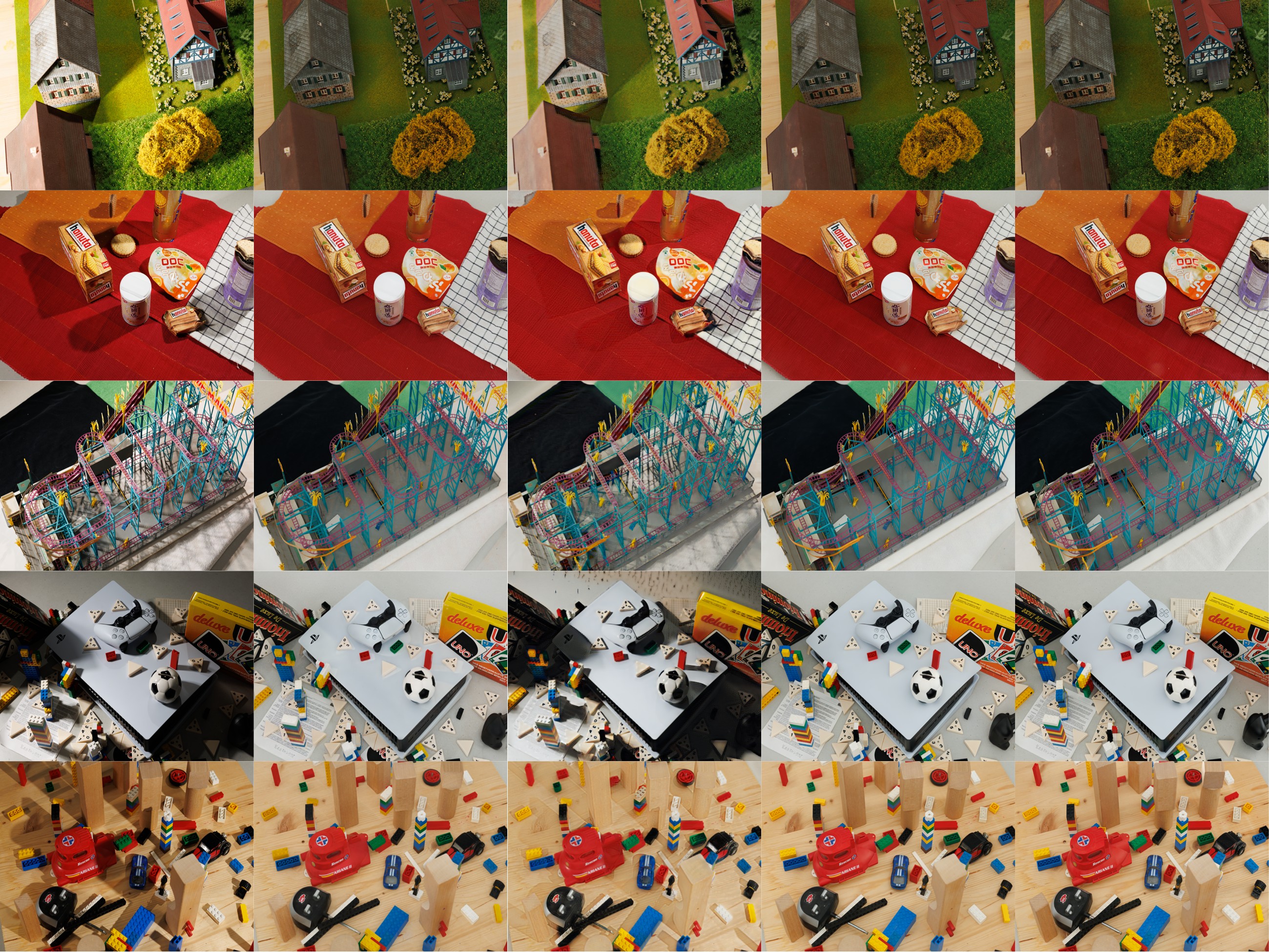}\\
    \begin{minipage}[c]{0.16\textwidth}
    \centering
    \scriptsize{Input}
    \end{minipage}
    \begin{minipage}[c]{0.16\textwidth}
    \centering
    \scriptsize{PromptNorm\cite{Serrano-Lozano_2025_CVPR}}
    \end{minipage}
    \begin{minipage}[c]{0.16\textwidth}
    \centering
    \scriptsize{DCShadowNet~\cite{jin2021dcshadownet}}
    \end{minipage}
    \begin{minipage}[c]{0.16\textwidth}
    \centering
   \scriptsize{\textbf{Ours}}
    \end{minipage}
    \begin{minipage}[c]{0.16\textwidth}
    \centering
   \scriptsize{\textbf{GT}}
    \end{minipage}
    \caption{Qualitative comparisons on the Ambient6K dataset. From left to right, we show the input image, PromptNorm~\cite{Serrano-Lozano_2025_CVPR},  DCShadowNet~\cite{jin2021dcshadownet}, our UniBlendNet, and the ground truth. Compared with the competing methods, UniBlendNet produces more consistent illumination normalization and cleaner restoration in challenging shadow regions.}
    \label{fig:vis_compare1}
\end{figure*}

\noindent \textbf{Dataset.}
We evaluate the proposed UniBlendNet on the \textbf{Ambient6K} dataset~\cite{ifblend,vasluianu2025ntirealn}, which is designed for Ambient Lighting Normalization under complex illumination conditions. Compared with conventional shadow removal benchmarks such as SRD~\cite{qu2017deshadownet} and ISTD~\cite{wang2018stackedgan}, as well as the more recent WSRD benchmark~\cite{vasluianu2023wsrd}, Ambient6K contains more realistic scenes with multiple light sources, cast shadows, self-shadows, and diverse material properties, making it more challenging for illumination restoration. Following the standard setting~\cite{ifblend,vasluianu2025ntirealn}, we use the official training-testing split for training and evaluation.

\noindent \textbf{Implementation Details and Metrics.}
Our method is implemented in PyTorch~\cite{paszke2019pytorch} and trained on RTX 4070 and 2080Ti GPUs. For data augmentation, we adopt random cropping with a patch size of \textbf{384 $\times$ 384}, together with vertical and horizontal flipping. We use the Adam optimizer~\cite{kingma2015adam} for training, where $\beta_1$ and $\beta_2$ are set to $0.9$ and $0.999$, respectively. The initial learning rate is set to $1 \times 10^{-4}$, and the model is trained for 100 epochs. During training, we adopt the multi-objective loss defined in Eq.~(\ref{eq:total_loss}), which combines reconstruction loss, SSIM loss, gradient loss, perceptual loss, and mask supervision loss. Inference is performed at full resolution without resizing. We adopt Peak Signal-to-Noise Ratio (PSNR), Structural Similarity Index Measure (SSIM)~\cite{wang2004ssim}, and Learned Perceptual Image Patch Similarity (LPIPS)~\cite{zhang2018lpips} for quantitative metrics.


\begin{figure*}[t]
    \centering
    \setlength{\abovecaptionskip}{1mm}
    \includegraphics[width=0.92\textwidth]{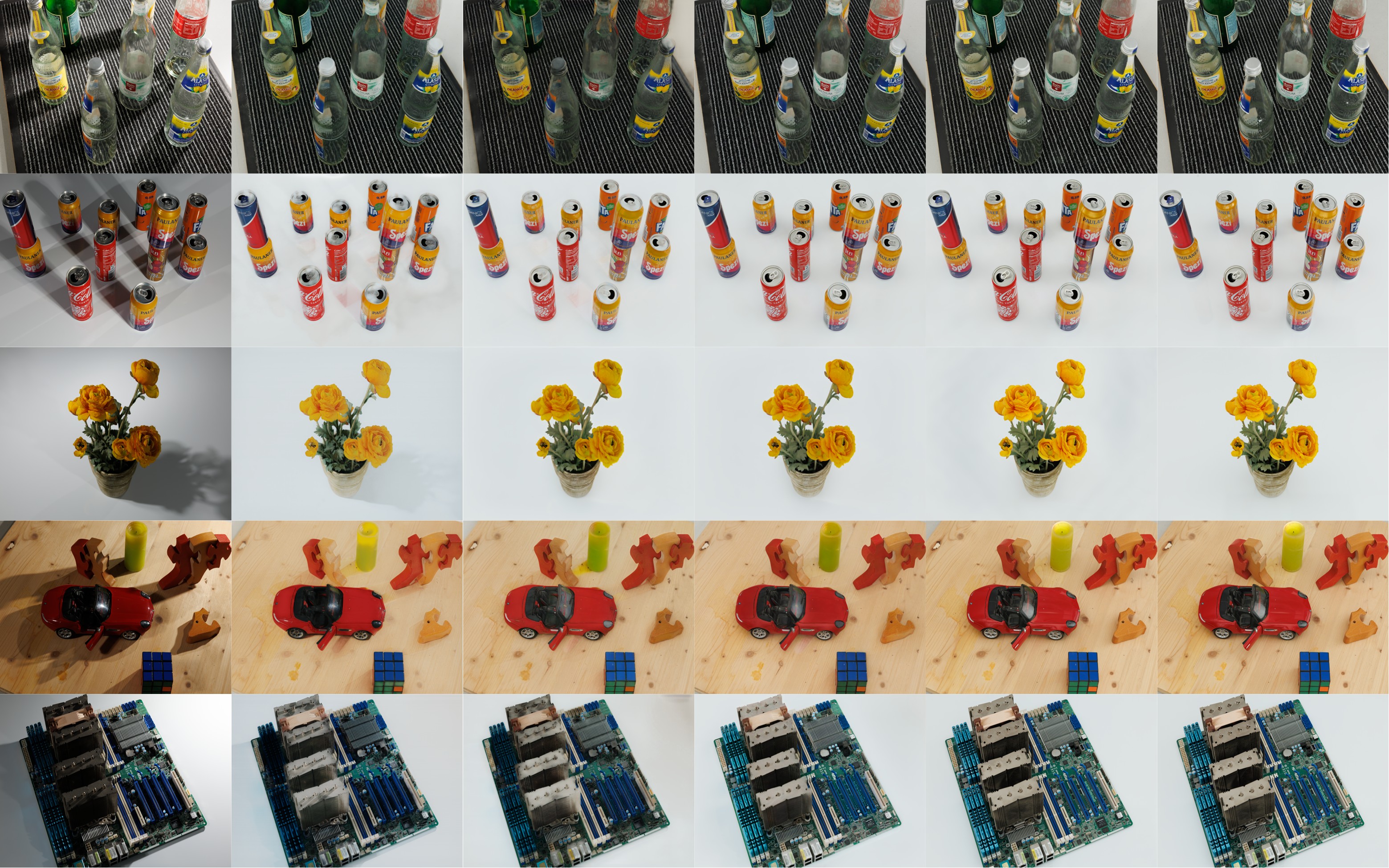}\\[0.0mm]

    {\scriptsize
    \makebox[0.145\textwidth][c]{Input}
    \hspace{0.012\textwidth}
    \makebox[0.145\textwidth][c]{Baseline}
    \hspace{0.000\textwidth}
    \makebox[0.145\textwidth][c]{Baseline + Mask}
    \hspace{-0.008\textwidth}
    \makebox[0.175\textwidth][c]{Baseline + Mask + SAAM}
    \hspace{-0.010\textwidth}
    \makebox[0.145\textwidth][c]{\textbf{Full model}}
    \hspace{0.010\textwidth}
    \makebox[0.120\textwidth][c]{\textbf{GT}}
    }

    \caption{Qualitative visualization of the ablation study on the Ambient6K dataset. From left to right, we show the input image, the IFBlend baseline, the baseline with mask-guided residual refinement, the baseline further enhanced with multi-scale aggregation, the full UniBlendNet, and the ground truth. The progressive results show that the proposed components improve restoration quality in a complementary manner, leading to better illumination consistency, fewer over-enhancement artifacts, and more faithful recovery in severely degraded regions.}
    \label{fig:vis_compare2}
    \vspace{-2mm}
\end{figure*}

\subsection{Comparison with Representative Methods}

We compare the proposed UniBlendNet with several representative methods, including mask-free methods \textbf{PromptNorm}~\cite{Serrano-Lozano_2025_CVPR}, \textbf{DCShadowNet}~\cite{jin2021dcshadownet}, and the \textbf{IFBlend} baseline~\cite{ifblend}, as well as mask-based methods \textbf{ShadowFormer}~\cite{guo2023shadowformer}. These baselines are selected to cover both conventional shadow removal frameworks and recent high-performance restoration architectures. Since PromptNorm is a recent method specifically designed for ambient lighting normalization, we replace Refusion with PromptNorm~\cite{Serrano-Lozano_2025_CVPR} in the main comparison. For fair comparison, all compared methods were retrained on the Ambient6K training split under the same data setting.


\noindent \textbf{Quantitative Comparison.}
The quantitative comparison on the Ambient6K validation set is reported in Tab.~\ref{tab:main_compare}. Overall, UniBlendNet achieves the best performance across all three metrics, reaching \textbf{25.237} PSNR, \textbf{0.864} SSIM, and \textbf{0.083} LPIPS. Compared with the strongest baseline PromptNorm, our method improves PSNR by 3.418 dB, increases SSIM by 0.043, and reduces LPIPS by 0.064, demonstrating the effectiveness of our global context modeling, scale-aware multi-scale aggregation, and mask-guided residual refinement under complex illuminations.

\begin{table}[t]
\setlength{\abovecaptionskip}{1mm}
\centering
\begin{tabular}{lccc}
\hline
Method & PSNR$\uparrow$ & SSIM$\uparrow$ & LPIPS$\downarrow$ \\
\hline
DCShadowNet~\cite{jin2021dcshadownet} & 19.961 & 0.799 & 0.179 \\
ShadowFormer~\cite{guo2023shadowformer} & 20.106 & 0.815 & 0.176 \\
IFBlend~\cite{ifblend} & 20.314 & 0.783 & 0.165 \\
PromptNorm\cite{Serrano-Lozano_2025_CVPR} & 21.819 & 0.821 & 0.147 \\
\textbf{UniBlendNet (ours)} & \textbf{25.237} & \textbf{0.864} & \textbf{0.083} \\
\hline
\end{tabular}
\caption{Quantitative comparison with representative methods on the Ambient6K validation set. Best results are highlighted in bold.}
\label{tab:main_compare}
\vspace{-2mm}
\end{table}

\noindent \textbf{Visual Comparison.}
We provide qualitative comparisons with representative competing methods in Fig.~\ref{fig:vis_compare1}. Compared with other baselines, UniBlendNet produces more consistent illumination normalization and better preserves object structure and color fidelity in shadow regions. In particular, our method is more effective in handling difficult cases with strong shadow boundaries, non-uniform illumination transitions, and complex local textures.

\subsection{Comparison on the Official Test Set}

To further verify the effectiveness of the proposed method under the official evaluation protocol~\cite{vasluianu2025ntirealn}, we compare UniBlendNet with IFBlend on the official test set and report quantitative results in Tab.~\ref{tab:official_test_compare}. UniBlendNet achieves clear improvements over IFBlend in both PSNR and SSIM, demonstrating the effectiveness of the introduced UniConvNet-based global context modeling, Scale-Aware Aggregation Module, and mask-guided residual refinement. Compared with IFBlend, UniBlendNet improves PSNR by 4.529 dB and SSIM by 0.0773. These improvements verify that enhancing IFBlend with stronger global context modeling, scale-aware feature aggregation, and spatially adaptive refinement can significantly improve restoration quality under complex non-uniform illumination conditions.

\begin{table}[t]
\setlength{\abovecaptionskip}{1mm}
\centering
\begin{tabular}{lccc}
\hline
Method & PSNR$\uparrow$ & SSIM$\uparrow$ & LPIPS$\downarrow$ \\
\hline
IFBlend~\cite{ifblend} & 20.213 & 0.7732 & 0.1896 \\
\textbf{UniBlendNet (ours)} & \textbf{24.742} & \textbf{0.8505} & \textbf{0.1052} \\
\hline
\end{tabular}
\caption{Quantitative comparison with the IFBlend baseline on the official test set. Best results are highlighted in bold.}
\label{tab:official_test_compare}
\vspace{-8mm}
\end{table}

\subsection{Ablation Study}

To verify the contribution of each component, we conduct ablation studies on the Ambient6K validation set. Specifically, we analyze the effects of the mask-guided residual refinement mechanism, Scale-Aware Aggregation Module (SAAM), and the UniConvNet-based global context branch. The quantitative and qualitative results are summarized in Tab.~\ref{tab:ablation_compare} and Fig.~\ref{fig:vis_compare2}, respectively. Starting from the IFBlend baseline, introducing the mask-guided refinement improves PSNR from 20.314 to 22.036 and SSIM from 0.783 to 0.806, while reducing LPIPS from 0.165 to 0.129. Adding SAAM further improves the performance to 23.926 PSNR, 0.833 SSIM, and 0.114 LPIPS. Finally, incorporating the UniConvNet-based global context branch yields the full UniBlendNet and achieves the best overall result, reaching \textbf{25.237} PSNR, \textbf{0.864} SSIM, and \textbf{0.083} LPIPS.

\begin{table}[t]
\setlength{\abovecaptionskip}{1mm}
\centering
\begin{tabular}{lccc}
\hline
Configuration & PSNR$\uparrow$ & SSIM$\uparrow$ & LPIPS$\downarrow$ \\
\hline
Baseline & 20.314 & 0.783 & 0.165 \\
Baseline + Mask & 22.036 & 0.806 & 0.129 \\
Baseline + Mask + SAAM & 23.926 & 0.833 & 0.114 \\
Full model & \textbf{25.237} & \textbf{0.864} & \textbf{0.083} \\
\hline
\end{tabular}
\caption{Ablation study on the Ambient6K validation set.}
\label{tab:ablation_compare}
\vspace{-2mm}
\end{table}

\begin{table}[t]
\setlength{\abovecaptionskip}{1mm}
\centering
\begin{tabular}{lccc}
\hline
Configuration & Params$\downarrow$ & GMACs$\downarrow$ & Time$\downarrow$ \\
\hline
Baseline & 272.0M & 1245.1 & 227.9 ms \\
+ Mask & 272.0M & 1245.4 & 230.3 ms \\
+ Mask\&SAAM & 277.7M & 1250.4 & 229.2 ms \\
Full model & 265.2M & 1000.1 & 334.6 ms \\
\hline
\end{tabular}
\caption{Complexity comparison of different model variants.}
\label{tab:complexity_compare}
\vspace{-2mm}
\end{table}

\noindent \textbf{Complexity Analysis.}
Tab.~\ref{tab:complexity_compare} reports the computational complexity of different model variants. Compared with the IFBlend baseline, the full model reduces the parameter count from 272.0M to 265.2M and the computational cost from 1245.1 GMACs to 1000.1 GMACs, while the inference time increases from 227.9 ms to 334.6 ms. Therefore, the proposed model is lighter in terms of parameter count and theoretical computation, but slower in wall-clock inference time. In return, it provides the strongest restoration performance among all ablation variants.

\noindent\textbf{Effect of mask-guided refinement.}
As shown in Fig.~\ref{fig:vis_compare2}, the mask-guided residual mechanism explicitly introduces spatial adaptivity into the restoration process. Compared with uniform residual correction, it allows the model to focus more on degraded shadow regions while preserving well-exposed areas, thereby reducing over-enhancement artifacts. In terms of complexity, this module introduces only marginal overhead over the baseline.

\noindent\textbf{Effect of scale-aware aggregation.}
Introducing SAAM enables the network to aggregate features from multiple spatial scales adaptively. This improves robustness to different shadow sizes and illumination variations, especially in images containing both smooth shading transitions and local structural details, as also reflected by the progressive visual improvements in Fig.~\ref{fig:vis_compare2}. The complexity increase brought by this module is modest compared with the baseline.

\noindent\textbf{Effect of global context enhancement.}
Incorporating the UniConvNet-based global context branch yields the full model and improves long-range dependency modeling and scene-level illumination understanding, leading to more stable correction under complex lighting patterns. Although the full model has fewer parameters and lower GMACs than the baseline, its runtime latency is higher in practice.

Overall, the ablation results show that the three proposed components are complementary. The mask-guided refinement introduces spatial adaptivity, SAAM improves multi-scale representation, and the UniConvNet branch strengthens global illumination modeling. Their combination leads to the best restoration performance on the Ambient6K validation set.
\section{Conclusion}

In this paper, we proposed UniBlendNet, an enhanced IFBlend-based framework for Ambient Lighting Normalization. The proposed method integrates a UniConvNet-based global context branch, a Scale-Aware Aggregation Module (SAAM), and a mask-guided residual refinement mechanism for global illumination modeling, multi-scale feature aggregation, and spatially adaptive restoration. Experiments on the NTIRE Ambient Lighting Normalization benchmark show that UniBlendNet consistently outperforms the baseline IFBlend in both quantitative metrics and visual quality. Ablation studies further confirm the effectiveness of all three components. Overall, the results verify that combining global context modeling, scale-aware aggregation, and guidance-mask-based refinement is effective for Ambient Lighting Normalization.

{
    \small
    \bibliographystyle{ieeenat_fullname}
    \bibliography{main}
}

\end{document}